\documentclass[runningheads]{llncs}
\usepackage[T1]{fontenc}
\usepackage{graphicx,verbatim}
\usepackage{amsmath}
\usepackage{amssymb}
\usepackage{color}
\usepackage{url}
\usepackage{hyperref}

\usepackage{booktabs}
\usepackage{multirow}
\usepackage{siunitx}
\usepackage{tabularx}

\begin{document}
\title{WING: A Window-Prior-Based Generative Network with Gated Inception for Cross-Modality CT Synthesis}
\titlerunning{WING: A Window-Prior-Based Generative Network for CT Synthesis}
\begin{comment}  %% Removed for anonymized MICCAI submission
\author{First Author\inst{1}\orcidID{0000-1111-2222-3333} \and
Second Author\inst{2,3}\orcidID{1111-2222-3333-4444} \and
Third Author\inst{3}\orcidID{2222--3333-4444-5555}}
%
\authorrunning{F. Author et al.}
\institute{Princeton University, Princeton NJ 08544, USA \and
Springer Heidelberg, Tiergartenstr. 17, 69121 Heidelberg, Germany
\email{lncs@springer.com}\\
\url{http://www.springer.com/gp/computer-science/lncs} \and
ABC Institute, Rupert-Karls-University Heidelberg, Heidelberg, Germany\\
\email{\{abc,lncs\}@uni-heidelberg.de}}

\end{comment}

% \author{Anonymized Authors}  %% Added for anonymized MICCAI submission
% \authorrunning{Anonymized Author et al.}
% \institute{Anonymized Affiliations \\
%     \email{email@anonymized.com}}

\author{Siyuan Mei\inst{1}
\and
Yan Xia\inst{2} \and
Yipeng Sun\inst{1} \and
Siming Bayer\inst{1} \and
Zirong Li\inst{3} \and
Chengze Ye\inst{1} \and
Daiqi Liu\inst{1} \and
Fuxin Fan\inst{3} \and
Yixing Huang\inst{4} \and
Andreas Maier\inst{1}}
\authorrunning{S. Mei et al.}
\institute{
Pattern Recognition Lab, Friedrich-Alexander-Universit\"at Erlangen-N\"urnberg, Erlangen 91058, Germany\\
\email{siyuan.mei@fau.de}
\and
Department of Orthodontics and Orofacial Orthopaedics, Friedrich-Alexander-Universit\"at Erlangen-N\"urnberg, Erlangen 91054, Germany
\and
Digital Technology and Innovation, Siemens Healthineers, Shanghai 201318, China
\and
Institute of Medical Technology, Peking University, Beijing 100191, China
}
  
\maketitle              % typeset the header of the contribution
\begin{abstract}
Generating CT volumes from MRI and CBCT can improve treatment planning in adaptive radiotherapy while avoiding additional radiation exposure. However, direct regression of CT intensities is challenged by the inherently high dynamic range and long-tailed distributions, thereby averaging out sparse yet clinically important structures. To alleviate this issue, we reformulate the regression target into multiple windowed representations, leveraging the inductive prior that CT intensities are structure-deterministic and window-separable. These windowed views exhibit smoother distributions and admit structured fusion back to the full-range CT. Building on this reformulation, we introduce WING, a WINdow-prior-based Generative network comprising: 1) a new Gated Inception Generator to produce multi-window predictions, enabling multi-shape kernel interactions to capture cross-modality correspondence; 2) a Fuse-and-Refine Transformer to aggregate the windowed outputs and learn residuals for detail refinement; and 3) a joint adversarial training objective to enhance window-conditioned realism. Extensive experiments demonstrate that our compact WING achieves state-of-the-art performance on the MRI-to-CT and CBCT-to-CT benchmarks, while supporting multi-anatomy synthesis with a single model. 

% Authors must provide keywords and are not allowed to remove this Keyword section.

\end{abstract}

\section{Introduction}

\begin{figure}[htp]
\includegraphics[width=\textwidth]{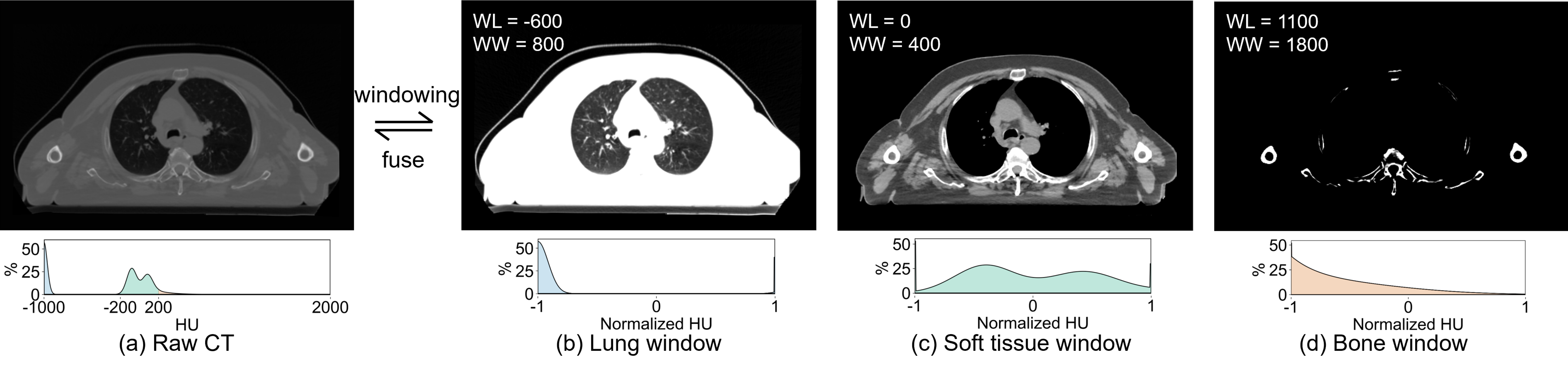}
\caption{Illustration of raw CT images and three non-overlapping windows of lung, soft tissue, and bone. The windowed views are obtained using the fixed window level (WL) and window width (WW), featuring more learnable distributions.} \label{fig1}
\end{figure}

Adaptive radiotherapy (ART) leverages repeated on-treatment imaging, most commonly cone-beam computed tomography (CBCT) on conventional linacs and magnetic resonance imaging (MRI) on MR-guided systems, to capture inter-fraction anatomical variability and update treatment plans~\cite{glide2021adaptive,thummerer2025synthrad2025,sonke2019adaptive}. While MRI provides excellent soft-tissue contrast and CBCT involves lower radiation exposure, neither modality directly yields CT-calibrated X-ray attenuation required for accurate proton and photon dose calculation~\cite{gregoire2011state,thummerer2021clinical}. 
To optimize treatment planning in MRI- or CBCT-only ART workflows, synthetic CT (sCT) has recently been introduced to provide CT-equivalent electron density information, thereby avoiding additional diagnostic CT acquisitions and associated patient radiation burden~\cite{sct1,thummerer2025synthrad2025,synthrad2023}.

% Consequently, CT remains the reference for electron-density information, creating a bottleneck for MRI- or CBCT-only ART workflows. Synthetic CT (sCT) aims to bridge this gap by predicting CT-driven density maps from MRI or CBCT, enabling CT-sparing dose computation and plan adaptation without additional diagnostic CT acquisitions~\cite{sct1,thummerer2025synthrad2025,synthrad2023}.

% Adaptive radiotherapy (ART) employs magnetic resonance imaging (MRI) or cone-beam computed tomography (CBCT) to frequently monitor anatomical changes and re-adjust radiation treatment plans throughout the course of therapy~\cite{glide2021adaptive,thummerer2025synthrad2025,sonke2019adaptive}. 
% MRI offers superior soft-tissue contrast, whereas CBCT involves lower radiation exposure, making both more suitable for repeated imaging than conventional CT~\cite{glide2021adaptive,synthrad2023}.
% % ~\cite{mri,mri2} determine beam attenuation and
% Nevertheless, conformal radiation treatment still relies on CT-derived electron density maps, which are required for accurate proton and photon dose calculation~\cite{gregoire2011state,thummerer2021clinical}. To optimize treatment planning in MRI- or CBCT-only ART workflows, synthetic CT (sCT) has recently been introduced to provide CT-free electron density information, thereby avoiding additional radiation burden to patients~\cite{sct1,thummerer2025synthrad2025,synthrad2023}.

A wide range of deep learning models have been applied to sCT generation by casting it as a 3D image-to-image translation problem~\cite{sct1,guo2025deep,synthrad2023}. 
Early approaches mainly rely on convolutional neural networks (CNNs), where U-Net~\cite{ronneberger2015u} and its variants~\cite{isensee2021nnu,yoganathan2023generating} are widely adopted due to their efficient encoder-decoder structures. 
To further improve image fidelity, subsequent studies introduce advanced generative paradigms, including adversarial generative models (GAN) with discriminators~\cite{pix2pix,yu2019eagan,zhu2017unpaired} and diffusion models with iterative denoising processes~\cite{peng2024cbct,gong2025boundary}. 
More recently, higher-capacity modern architectures such as Transformers~\cite{fuxin} and ConvNeXts~\cite{roy2023mednext,mei2025ganeXt} have also been explored to capture richer spatial representations with long-range dependencies.

Despite their architectural diversity, these methods overlook CT-specific imaging physics and intensity distributions. Unlike natural images with context-relative pixel values, CT intensities are absolute Hounsfield units (HU) derived from tissue-dependent X-ray attenuation. Consequently, CT images exhibit a substantially wider dynamic range (e.g., -1000 to 3000 HU) and highly imbalanced, long-tailed distributions. 
As illustrated in Fig.~\ref {fig1} (a), a single global intensity window often fails to simultaneously emphasize low-density regions (e.g., lung parenchyma and air cavities), soft tissues (e.g., tumors and organs-at-risk), and high-density bony structures used for robust spatial anchoring. 
Direct full-range regression causes learning to be dominated by abundant soft-tissue voxels ($[-200, 200]$~HU), thereby suppressing long-tail structures.
While some strategies introduce geometric guidance to recover degraded details~\cite{yu2019eagan,gong2025boundary,phan2024structural}, they do not fundamentally resolve this imbalance and often need external segmentation models or labels that are impractical in clinical settings~\cite{poch2025segmentation,phan2024structural}.

In contrast, CT images themselves establish complementary anatomical cues with clinically relevant contrasts through the windowing process~\cite{ma2025vision}. As shown in Fig.~\ref{fig1} (b-d), windowed representations preserve a deterministic correspondence between fixed intensity ranges and anatomical structures while exhibiting smoother data distributions. Based on this observation, we reformulate sCT generation by shifting the regression target from full-range CT to multiple windowed representations, explicitly utilizing the structure determinism and window separability of CT as inductive priors. Specifically, we propose WING, a WINdow-prior-based Generative network with the following core designs:
\begin{itemize}
\item We introduce a Gated Inception Generator (GIG) to generate multi-window predictions. Built upon the compact MedNeXt backbone~\cite{roy2023mednext}, we perform an Inception-inspired branch decomposition~\cite{yu2024inceptionnext} to replace the original large-kernel convolutions, which effectively improve efficiency and performance.
\item We propose a Fuse-and-Refine Transformer (FRT) that aggregates the synthesized windowed views through a differentiable soft fusion function and learns residuals for refinement of structural details.
\item We propose a joint adversarial training objective that simultaneously optimizes the windowed and fused results, ensuring dual-level perceptual enhancement.
\end{itemize}
Experimental results show that the proposed WING outperforms previous SoTA performance on both MRI- and CBCT-to-CT tasks, while enabling multi-anatomy synthesis with a single compact model.

% \section{Related Work}

% \subsection{Image-to-image Translation}

% CT generation from a paired conditional image is commonly formulated as a 3D image-to-image translation problem. 

% \subsection{Structural-guided CT Synthesis}

\section{Methods}

\subsection{CT Windowing and Soft Fusion}

\paragraph{CT Windowing:}
Given a CT volume $\mathbf{y}_{ct} \in \mathbb{R}^{D \times H \times W}$, CT windowing highlights desired structures using a predefined window level (WL) and window width (WW), followed by normalization to the range of $[-1,1]$:
\begin{equation}
\mathbf{y}_{(w)} =
\frac{
\mathrm{clip}\!\left(
\mathbf{y}_{ct},\;
\mathrm{WL} - \mathrm{WW}/{2},\;
\mathrm{WL} + \mathrm{WW}/{2}
\right)
- \mathrm{WL}
}{\mathrm{WW}/2}.
\end{equation}
Following radiological practice~\cite{murphy2017windowingct}, we adopt three non-overlapping windows corresponding to lung (WL: $-600$, WW: $800$), soft tissue (WL: $0$, WW: $400$), and bone (WL: $1100$, WW: $1800$) (see Fig.~\ref{fig1}). The windowed views are stacked to form a multi-channel representation:
\begin{equation}
\label{windowing}
\mathbf{y}_{win} =
\left[
\mathbf{y}_{(\mathrm{lung})},\;
\mathbf{y}_{(\mathrm{soft})},\;
\mathbf{y}_{(\mathrm{bone})}
\right]
\in \mathbb{R}^{3 \times D \times H \times W}.
\end{equation}

\paragraph{Soft Fusion:}

Fusion operation serves as an inverse reconstruction of the windowing process. To this end, each window is first denormalized to yield $\tilde{\mathbf{y}}_{(\mathrm{lung})}$,
$\tilde{\mathbf{y}}_{(\mathrm{soft})}$, and
$\tilde{\mathbf{y}}_{(\mathrm{bone})}$.
Rather than using hard fusion boundaries, we propose a \emph{differentiable} soft fusion function that assigns weighted masks to enable smooth blending across windows.
The fusion weights are derived from the medium (soft tissue) window, whose lower and upper bounds characterize the transitions between windows. Specifically, the lung and bone masks are defined as
\begin{equation}
\label{eq:soft_fuse_masks}
\mathbf{m}_{(\mathrm{lung})} =
\mathrm{clip}\!\left(
-\frac{\mathbf{y}_{(\mathrm{soft})} + \tau}{1 - \tau},\; 0,\; 1
\right), 
\quad
\mathbf{m}_{(\mathrm{bone})} =
\mathrm{clip}\!\left(
\frac{\mathbf{y}_{(\mathrm{soft})} - \tau}{1 - \tau},\; 0,\; 1
\right).
\end{equation}
The soft-tissue mask, defined as
$\mathbf{m}_{(\mathrm{soft})} = 1 - \mathbf{m}_{(\mathrm{lung})} - \mathbf{m}_{(\mathrm{bone})}$, forms a trapezoidal profile and takes the value of 1 for $\mathbf{y}_{(\mathrm{soft})} \in [-\tau, \tau]$.
The fused CT image is computed as the sum of weighted windows:
% \begin{equation}
% \label{soft fusion}
% \mathbf{y}_{\mathrm{fused}} =
% \mathbf{m}_{\mathrm{lung}} \cdot \tilde{\mathbf{y}}_{(\mathrm{lung})}
% + \mathbf{m}_{\mathrm{soft}} \cdot \tilde{\mathbf{y}}_{(\mathrm{soft})}
% + \mathbf{m}_{\mathrm{bone}} \cdot \tilde{\mathbf{y}}_{(\mathrm{bone})},
% \end{equation}
\begin{equation}
\label{eq:soft_fusion}
\mathbf{y}_{\mathrm{fused}}
=
\sum_{(w)}
\mathbf{m}_{(w)}\cdot\tilde{\mathbf{y}}_{(w)},\quad 
w\in\{\mathrm{lung},\,\mathrm{soft},\,\mathrm{bone}\}
\end{equation}
where $\tau \in [0, 1)$ controls the hardness of the fusion, with larger values resulting in sharper transitions between windows.
In this paper, we set $\tau = 0.2$ to obtain smooth and stable fusion boundaries.

\subsection{WING Model}

\begin{figure}[t]
\includegraphics[width=\textwidth]{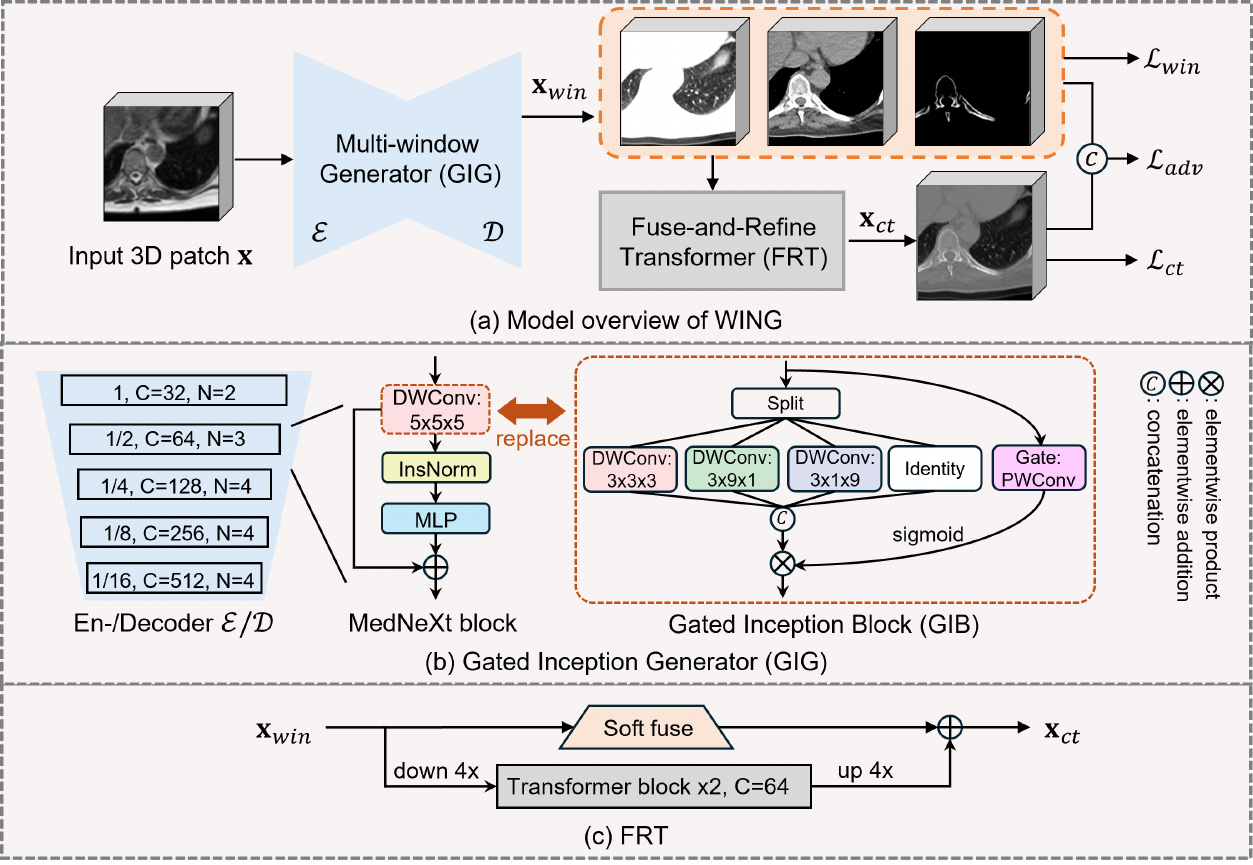}
\caption{Model overview of WING and block illustration of GIG and FRT. ``DWConv'' means depth-wise convolution~\cite{howard2017mobilenets}, and ``PWConv'' means point-wise convolution~\cite{howard2017mobilenets}. ``Soft fuse'' denotes a differentiable operator defined in Eq.~\ref{eq:soft_fusion}.} \label{fig2}
\end{figure}

We then introduce the proposed WING model, including the Gated Inception Generator (GIG), the Fuse-and-Refine Transformer (FRT), and the joint adversarial objective. An overview of the whole framework is illustrated in Fig.~\ref{fig2}.

\subsubsection{Gated Inception Generator}
We build our generator upon the MedNeXt backbone for its compact design; architectural details are referred to~\cite{roy2023mednext}. 
As illustrated in Fig.~\ref{fig2}(b), the backbone follows a symmetric hierarchical encoder-decoder net, where feature resolutions are progressively downsampled from $1$ to $1/16$, channel dimensions $C$ increase accordingly, and $N$ blocks are stacked at each stage. As its core, MedNeXt employs a residual block composed of a large-kernel 3D depthwise convolution (DWConv)~\cite{howard2017mobilenets}, followed by an instance normalization layer and a multi-layer perceptron (MLP) layer. 

While large-kernel DWConv effectively enlarges the receptive field, its computational cost grows cubically with kernel size, which becomes prohibitive for volumetric inputs. To address this limitation, we replace the original DWConv with a multi-branch \emph{Gated Inception Block} (GIB) inspired by~\cite{yu2024inceptionnext}. As shown on the right side of Fig.~\ref{fig2}(b), input channels are evenly split into four branches, processed by DWConv layers with kernels $3\!\times\!3\!\times\!3$, $3\!\times\!9\!\times\!1$, and $3\!\times\!1\!\times\!9$, along with an identity branch. All four branches are computationally much more efficient than a large-kernel DWConv, and they can work together to aggregate multi-directional spatial information. In particular, the anisotropic kernels expand the in-plane receptive field while keeping local through-plane mixing, helping preserve thin, high-contrast structures without inducing inter-slice blurring. 

To integrate branch-wise features, we adopt a gated fusion via element-wise modulation. Specifically, a lightweight point-wise convolution (PWConv)~\cite{howard2017mobilenets} followed by Sigmoid produces a gate map that adaptively reweights branch responses. Compared to direct concatenation, this mechanism implements a learnable mixture of multiple branches, yielding more structure-aware features. Finally, the decoder outputs are projected with a Tanh activation to produce three channels corresponding to lung, soft-tissue, and bone window predictions $\mathbf{x}_{win}$.

% This design reduces computational complexity while capturing anisotropic spatial context, which is well aligned with the geometry of 3D medical images.

% To integrate branch-wise features, we introduce a gated fusion mechanism based on element-wise modulation. The gate map is generated via a lightweight point-wise convolution (PWConv)~\cite{howard2017mobilenets} followed by a Sigmoid activation, enabling adaptive reweighting of branch responses. Compared to direct concatenation, this design introduces nonlinearity into the fusion process and enables adaptive reweighting of branch-specific responses, yielding more expressive and structure-aware representations. Finally, the decoded features through the whole generator are projected with a Tanh activation to produce a three-channel output, corresponding to the lung, soft-tissue, and bone window predictions $\mathbf{x}_{win}$.

\subsubsection{Fuse-and-Refine Transformer}
Since a gap exists between the predicted windows and the real ones, direct fusion is insufficient to reconstruct a faithful full-range CT. Therefore, FRT introduces a core component, Transformer~\cite{vaswani2017attention}, to refine the coarse fusion in a residual manner.

As illustrated in Fig.~\ref{fig2}(c), the windowed predictions $\mathbf{x}_{win}$ are fed in parallel into the soft fusion operator (as depicted in Eq.~\ref{eq:soft_fusion}) and the Transformer path, producing a coarse CT estimate and its residual, respectively. For computational efficiency, $\mathbf{x}_{win}$ is first projected into $64$-channel patch embeddings at a spatial resolution reduced by a factor of $4$. The patch embeddings are then processed by two Transformer blocks, where self-attention is used to capture global dependencies across windowed representations. The Transformer output is subsequently upsampled and projected using two transposed convolution layers to restore the original spatial resolution. Finally, the predicted residual is added to the coarse fused CT, yielding the refined sCT $\mathbf{x}_{ct}$.

\subsubsection{Joint Adversarial Objective}
\label{loss}

To further enhance the realism of the synthesized results, we adopt a joint adversarial learning objective based on the PatchGAN framework~\cite{pix2pix}. 
Particularly, the joint discriminator (JDisc) $D(\cdot)$ is conditioned on windowed and full-range CT, enabling synchronized improvement of these coupled representations. The adversarial losses are defined as
\begin{equation}
\mathcal{L}_{\mathrm{adv}}^{D} = 0.5\cdot
\mathbb{E}_{\mathbf{y}}\!\left[(D([\mathbf{y}_{win}, \mathbf{y}_{ct}]) - 1)^2\right]
+
0.5\cdot\mathbb{E}_{\mathbf{x}}\!\left[D([\mathbf{x}_{win}, \mathbf{x}_{ct}])^2\right],
\end{equation}
\begin{equation}
\mathcal{L}_{\mathrm{adv}}^{G} =
\mathbb{E}_{\mathbf{x}}\!\left[(D([\mathbf{x}_{win}, \mathbf{x}_{ct}]) - 1)^2\right].
\end{equation}

For the reconstruction part, we apply $\ell_1$ losses to both the windowed outputs and the fused CT, together with an LPIPS loss~\cite{lpips} to improve perceptual quality:
\begin{equation}
\mathcal{L}_{\mathrm{rec}} =
\lambda_{\mathrm{win}}\|\mathbf{x}_{win} - \mathbf{y}_{win}\|_1
+
\lambda_{\mathrm{ct}}\|\mathbf{x}_{ct} - \mathbf{y}_{ct}\|_1
+
\lambda_{\mathrm{lpips}}\mathcal{L}_{\mathrm{lpips}}(\mathbf{x}_{ct}, \mathbf{y}_{ct}).
\label{eq7}
\end{equation}
The overall training objective of the generator is
\begin{equation}
\mathcal{L}_{G} =
\mathcal{L}_{\mathrm{rec}}
+
\lambda_{\mathrm{adv}} \mathcal{L}_{\mathrm{adv}}^{G},
\label{eq8}
\end{equation}
where $\lambda_{\mathrm{adv}}$ balances reconstruction accuracy and adversarial strength.

\section{Experiments}

\subsection{Dataset and Preprocessing}

We benchmark our models on the large-scale public SynthRAD2025~\cite{thummerer2025synthrad2025} challenge dataset, which contains around 600 paired MRI-to-CT and 950 CBCT-to-CT volumes from head-and-neck (HN), thoracic (TH), and abdominal (AB) cancer patients across five clinical centers. The dataset is further divided into a 7{:}1{:}2 split for training, validation, and testing. Unless specified, we jointly train on all three anatomical regions to promote robust cross-anatomy generalization.

For preprocessing, MRI and CBCT inputs are normalized to $[-1,1]$ using their 1\% and 99\% intensity percentiles, respectively. CT volumes are linearly scaled to $[-1,1]$ using fixed bounds of $-1024$ and $2000$ HU. Multi-window CT targets are then constructed following Eq.~\ref{windowing}. Since the original dataset provides only rigid alignment, we further conduct deformable registration on CT using ConvexAdam~\cite{siebert2024convexadam}, enabling accurate voxel-wise supervision.

\subsection{Implementation Details}
We implement the proposed WING using Pytorch~\cite{paszke2019pytorch}. The loss weights specified in Eq.~\ref{eq7} and Eq.~\ref{eq8} are empirically configured as: 
% $\lambda_{win}:\lambda_{ct}:\lambda_{lpips}:\lambda_{adv} = 1:3:0.4:0.08$.
\begin{equation*} \lambda_{\mathrm{win}}:\lambda_{\mathrm{ct}}:\lambda_{\mathrm{lpips}}:\lambda_{\mathrm{adv}}=1:3:0.4:0.1.
\end{equation*} 
Furthermore, we set a 3D patch size with the shape of $32 \times 160 \times 160$. During training, the generator uses the AdamW optimizer with a base learning rate of $5e-4$, a weight decay of $1e-2$, and a cosine decay scheduler; the discriminator uses the Adam optimizer with a base learning rate of $1e-3$. All experiments are conducted on an NVIDIA H100 GPU with a batch size of 2, training for a total of 1000 epochs, requiring around 20 hours. For data augmentation, we randomly sample two 3D patches from each volume and employ a random flip along the frontal axis. During inference, we adopt a sliding-window strategy with an overlap ratio of 0.7.

\begin{table}[!ht]
\centering
\caption{Quantitative results on MRI-to-CT and CBCT-to-CT tasks.}
\label{tab:main_results}
\setlength{\tabcolsep}{1pt}
\fontsize{8}{8}\selectfont
\resizebox{\linewidth}{!}{%
\begin{tabular}{l c cccc cccc}
\toprule
\multicolumn{10}{c}{\textbf{(a) System-level comparison}} \\
\addlinespace[2pt]
\multirow{2}{*}{Methods} & {\scriptsize Generator} &
\multicolumn{4}{c}{Task 1: MRI-to-CT} &
\multicolumn{4}{c}{Task 2: CBCT-to-CT} \\
\noalign{\vskip-2.0pt}
\cmidrule(lr){3-6}\cmidrule(lr){7-10}
& {\scriptsize \#Params(M)} &
MAE(HU)$\downarrow$ & MS-SSIM$\uparrow$ & PSNR$\uparrow$ & DICE$\uparrow$
& MAE(HU)$\downarrow$ & MS-SSIM$\uparrow$ & PSNR$\uparrow$ & DICE$\uparrow$ \\
\midrule

\multirow{2}{*}{Pix2Pix3D~\cite{pix2pix}} & \multirow{2}{*}{23.77} &
66.10 & 93.14 & 29.32 & 76.70 &
53.31 & 95.43 & 31.29 & 84.87 \\
& &
{\scriptsize $\pm$23.79} & {\scriptsize $\pm$4.31} & {\scriptsize $\pm$2.63} & {\scriptsize $\pm$7.56} &
{\scriptsize $\pm$14.44} & {\scriptsize $\pm$3.13} & {\scriptsize $\pm$2.43} & {\scriptsize $\pm$6.64} \\
\addlinespace[2pt]

\multirow{2}{*}{Ea-GAN~\cite{yu2019eagan}} & \multirow{2}{*}{56.50} &
64.79 & 93.32 & 29.42 & 77.81 &
51.70 & 95.64 & 31.46 & 85.54 \\
& &
{\scriptsize $\pm$23.98} & {\scriptsize $\pm$4.47} & {\scriptsize $\pm$2.67} & {\scriptsize $\pm$7.55} &
{\scriptsize $\pm$14.27} & {\scriptsize $\pm$3.26} & {\scriptsize $\pm$2.45} & {\scriptsize $\pm$6.31} \\
\addlinespace[2pt]

\multirow{2}{*}{ResUNet~\cite{isensee2021nnu}} & \multirow{2}{*}{102.32} &
66.30 & 93.23 & 29.31 & 76.73 &
52.97 & 95.55 & 31.31 & 85.18 \\
& &
{\scriptsize $\pm$23.71} & {\scriptsize $\pm$4.44} & {\scriptsize $\pm$2.63} & {\scriptsize $\pm$7.63} &
{\scriptsize $\pm$14.45} & {\scriptsize $\pm$3.36} & {\scriptsize $\pm$2.42} & {\scriptsize $\pm$6.34} \\
\addlinespace[2pt]

\multirow{2}{*}{SwinUNETR~\cite{hatamizadeh2021swin}} & \multirow{2}{*}{62.19} &
66.57 & 93.07 & \underline{29.67} & 75.32 &
53.46 & 95.55 & 31.27 & 85.13 \\
& &
{\scriptsize $\pm$24.90} & {\scriptsize $\pm$4.61} & {\scriptsize $\pm$2.62} & {\scriptsize $\pm$8.02} &
{\scriptsize $\pm$14.56} & {\scriptsize $\pm$3.27} & {\scriptsize $\pm$2.45} & {\scriptsize $\pm$6.42} \\
\addlinespace[2pt]

\multirow{2}{*}{MedNeXt~\cite{roy2023mednext}} & \multirow{2}{*}{17.47} &
\underline{63.50} & \underline{93.43} & 29.49 & \underline{78.80} &
\underline{50.57} & \underline{95.66} & \underline{31.56} & \underline{85.88} \\
& &
{\scriptsize $\pm$23.53} & {\scriptsize $\pm$4.49} & {\scriptsize $\pm$2.72} & {\scriptsize $\pm$7.05} &
{\scriptsize $\pm$14.61} & {\scriptsize $\pm$3.26} & {\scriptsize $\pm$2.50} & {\scriptsize $\pm$6.01} \\
\addlinespace[2pt]

\multirow{2}{*}{WING (Ours)} & \multirow{2}{*}{19.29} &
\textbf{61.92} & \textbf{93.94} & \textbf{29.89} & \textbf{79.11} &
\textbf{48.58} & \textbf{96.28} & \textbf{31.82} & \textbf{86.37} \\
& &
{\scriptsize $\pm$20.77} & {\scriptsize $\pm$3.55} &
{\scriptsize $\pm$2.49} & {\scriptsize $\pm$7.01} &
{\scriptsize $\pm$13.65} & {\scriptsize $\pm$3.21} &
{\scriptsize $\pm$2.49} & {\scriptsize $\pm$5.99} \\
\end{tabular}%
}

\setlength{\tabcolsep}{1pt}
\fontsize{9}{9}\selectfont
\resizebox{\linewidth}{!}{%
\begin{tabular}{l cccc cccc}
\toprule
\multicolumn{9}{c}{\textbf{(b) Per-window MAE comparison w. SoTA}} \\
\cmidrule(lr){2-5}\cmidrule(lr){6-9}
Methods & Lung & Soft tissue & Bone & Average & Lung & Soft tissue & Bone & Average \\
\midrule
% \cmidrule(lr){1-1}\cmidrule(lr){2-5}\cmidrule(lr){6-9}
MedNeXt~\cite{roy2023mednext}
& 23.17{\scriptsize$\pm$32.91} & 42.11{\scriptsize$\pm$17.13} & 178.86{\scriptsize$\pm$50.03} & 81.38{\scriptsize$\pm$22.71}
& 8.71{\scriptsize$\pm$4.82} & 37.18{\scriptsize$\pm$10.34} & 110.50{\scriptsize$\pm$28.44} & 52.13{\scriptsize$\pm$12.05} \\
\addlinespace[2pt]

WING (Ours)
& \textbf{17.39}{\scriptsize$\pm$21.76} & \textbf{41.80}{\scriptsize$\pm$16.98} & \textbf{167.16}{\scriptsize$\pm$48.93} & \textbf{75.45}{\scriptsize$\pm$6.18}
& \textbf{8.25}{\scriptsize$\pm$4.01} & \textbf{36.57}{\scriptsize$\pm$9.86} & \textbf{105.01}{\scriptsize$\pm$28.64} & \textbf{49.94}{\scriptsize$\pm$11.89} \\
\bottomrule
\end{tabular}%
}
\end{table}

\begin{figure}[!ht]
    \centering
    \includegraphics[width=\linewidth]{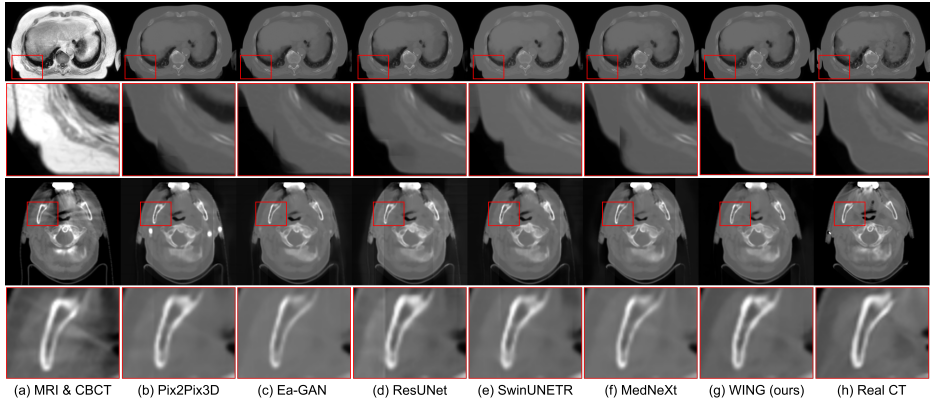}
    \caption{Visual comparison on MRI-to-CT and CBCT-to-CT tasks. A ×4 zoomed region is highlighted below. All gray values are windowed between -1000 HU and 1500 HU.}
    \label{fig:placeholder}
\end{figure}

\subsection{Performance Comparison}

% This systematic advancement stems from the mechanism of window regression. As shown in Fig.~\ref{fig:placeholder}, WING can produce realistic CT images across multiple modalities and anatomies. The highlighted regions further demonstrate that our approach preserves more accurate structural details (second row (g)) and exhibits reduced artifacts (fourth row (g)) compared to existing methods.

We compare WING with representative GANs w/wo geometric guidance, as well as the top three models in the SynthRAD challenge. Following~\cite {thummerer2025synthrad2025}, we evaluate image fidelity using MAE, MS-SSIM, and PSNR, and assess geometric consistency via DICE computed with a fixed pre-trained segmentation model. As shown in Table~\ref{tab:main_results} (a), WING achieves the best overall performance on both MRI-to-CT and CBCT-to-CT, while remaining parameter-efficient. Notably, our method surpasses the previous SoTA MedNeXt~\cite{roy2023mednext} by a significant margin: $-1.58$ MAE, $+0.51$ MS-SSIM, $+0.40$ PSNR, $+0.31$ DICE on MRI-to-CT, and $-1.99$ MAE, $+0.62$ MS-SSIM, $+0.26$ PSNR, $+0.49$ DICE on CBCT-to-CT.

% WING achieves the best overall performance on both MRI-to-CT and CBCT-to-CT and surpasses the previous SoTA MedNeXt~\cite{roy2023mednext} by a significant margin, while remaining parameter-efficient. 

Beyond system-level metrics, Table~\ref{tab:main_results}(b) reports per-window MAE under three clinical window settings, where errors are computed over voxels within the corresponding clipped intensity ranges. WING consistently outperforms the SoTA baseline across all windows and both tasks, with the largest gains observed on lung and bone structures that reside at the long-tailed extremes of the CT intensity distribution. These results indicate that the proposed window regression improves fidelity across heterogeneous attenuation regimes rather than merely optimizing global statistics. Visually, Fig.~\ref{fig:placeholder} demonstrates that WING can produce realistic CT images across multiple modalities and anatomies, exhibiting more accurate structural details (second row (g)) and reduced artifacts (fourth row (g)) compared to existing methods.

% consistent with the improved structural preservation and reduced artifacts observed in Fig.~\ref{fig:placeholder}.

\subsection{Ablation Study}

\begin{table}[t]
\centering
\caption{Ablation study of the proposed components on two tasks.}
\label{tab:ablation}
\setlength{\tabcolsep}{1pt}
\fontsize{8}{8}\selectfont
\resizebox{\linewidth}{!}{%
\begin{tabular}{l c cccc cccc}
\toprule
\multirow{2}{*}{Component} & \multirow{2}{*}{\#P(M)} &
\multicolumn{4}{c}{Task 1: MRI-to-CT} &
\multicolumn{4}{c}{Task 2: CBCT-to-CT} \\
\cmidrule(lr){3-6}\cmidrule(lr){7-10}
& &
MAE(HU)$\downarrow$ & MS-SSIM$\uparrow$ & PSNR$\uparrow$ & DICE$\uparrow$
& MAE(HU)$\downarrow$ & MS-SSIM$\uparrow$ & PSNR$\uparrow$ & DICE$\uparrow$ \\
\midrule

Baseline & \multirow{2}{*}{17.47} &
63.50 & 93.43 & 29.49 & 78.80 &
50.57 & 95.66 & 31.56 & 85.88 \\
(MedNeXt) & &
{\scriptsize $\pm$23.53} & {\scriptsize $\pm$4.49} & {\scriptsize $\pm$2.72} & {\scriptsize $\pm$7.05} &
{\scriptsize $\pm$14.61} & {\scriptsize $\pm$3.26} & {\scriptsize $\pm$2.50} & {\scriptsize $\pm$6.01} \\
\addlinespace[2pt]

\multirow{2}{*}{+ GIB} & \multirow{2}{*}{19.17} &
62.87 & 93.69 & 29.56 & 78.65 &
50.52 & 95.71 & 31.58 & 85.99 \\
& &
{\scriptsize $\pm$23.25} & {\scriptsize $\pm$4.30} & {\scriptsize $\pm$2.68} & {\scriptsize $\pm$7.25} &
{\scriptsize $\pm$14.56} & {\scriptsize $\pm$3.27} & {\scriptsize $\pm$2.53} & {\scriptsize $\pm$6.19} \\
\addlinespace[2pt]

\multirow{2}{*}{+ FRT} & \multirow{2}{*}{19.29} &
62.08 & 93.92 & 29.82 & 78.69 &
49.02 & 96.00 & 31.81 & 85.80 \\
& &
{\scriptsize $\pm$21.48} & {\scriptsize $\pm$3.50} & {\scriptsize $\pm$2.62} & {\scriptsize $\pm$7.25} &
{\scriptsize $\pm$13.57} & {\scriptsize $\pm$3.04} & {\scriptsize $\pm$2.47} & {\scriptsize $\pm$6.59} \\
\addlinespace[2pt]

+ JDisc & \multirow{2}{*}{30.32} &
\textbf{61.92} & \textbf{93.94} & \textbf{29.89} & \textbf{79.11} &
\textbf{48.58} & \textbf{96.28} & \textbf{31.82} & \textbf{86.37} \\
(Ours) & &
{\scriptsize $\pm$20.77} & {\scriptsize $\pm$3.55} &
{\scriptsize $\pm$2.49} & {\scriptsize $\pm$7.01} &
{\scriptsize $\pm$13.65} & {\scriptsize $\pm$3.21} &
{\scriptsize $\pm$2.49} & {\scriptsize $\pm$5.99} \\

\bottomrule
\end{tabular}}
\end{table}

As shown in Table~\ref{tab:ablation}, we conduct ablation studies to verify the effectiveness of each proposed component, including GIB, FRT, and JDisc. Starting from the MedNeXt baseline, we progressively introduce each module and observe consistent performance improvements across multiple evaluation metrics. Replacing the original block with GIB leads to noticeable gains while maintaining a similar complexity, highlighting feature enhancement through multi-kernel interactions. Moreover, FRT brings critical improvements with minimal parameter overhead, proving the necessity of window fusion and refinement strategies for CT modeling. Finally, incorporating JDisc results in additional gains, particularly in terms of DICE, reflecting improved geometric consistency and perceptual quality.

\section{Conclusion}

We present WING, the first window-prior-based generative framework for cross-modality CT synthesis. By explicitly incorporating imaging-driven window priors, WING alleviates the challenge of regressing long-tailed CT intensity distributions without relying on external geometric guidance, leading to consistent improvements on both MRI-to-CT and CBCT-to-CT tasks. These results demonstrate that WING can generate synthetic CT images with high fidelity and stable intensity characteristics, highlighting its potential for reliable use in radiotherapy workflows.

Nevertheless, several limitations should be noted. The predefined window settings follow standard radiological practice and may limit flexibility under uncommon acquisition protocols, and the separation of generation and fusion stages introduces additional design complexity. Future work will focus on more unified and adaptive window modeling strategies to further improve efficiency and generalizability.

\bibliographystyle{splncs04}
\bibliography{references}

@article{synthrad2023,
  title={Generating synthetic computed tomography for radiotherapy: SynthRAD2023 challenge report},
  author={Huijben, Evi MC and Terpstra, Maarten L and Pai, Suraj and Thummerer, Adrian and Koopmans, Peter and Afonso, Manya and Van Eijnatten, Maureen and Gurney-Champion, Oliver and Chen, Zeli and Zhang, Yiwen and others},
  journal={Medical image analysis},
  volume={97},
  pages={103276},
  year={2024},
  publisher={Elsevier}
}

@article{glide2021adaptive,
  title={Adaptive radiation therapy (ART) strategies and technical considerations: a state of the ART review from NRG oncology},
  author={Glide-Hurst, Carri K and Lee, Percy and Yock, Adam D and Olsen, Jeffrey R and Cao, Minsong and Siddiqui, Farzan and Parker, William and Doemer, Anthony and Rong, Yi and Kishan, Amar U and others},
  journal={International Journal of Radiation Oncology* Biology* Physics},
  volume={109},
  number={4},
  pages={1054--1075},
  year={2021},
  publisher={Elsevier}
}

@article{thummerer2025synthrad2025,
  title={SynthRAD2025 Grand Challenge dataset: Generating synthetic CTs for radiotherapy from head to abdomen},
  author={Thummerer, Adrian and van der Bijl, Erik and Galapon, Arthur Jr and Kamp, Florian and Savenije, Mark and Muijs, Christina and Aluwini, Shafak and Steenbakkers, Roel JHM and Beuel, Stephanie and Intven, Martijn PW and others},
  journal={Medical physics},
  volume={52},
  number={7},
  pages={e17981},
  year={2025},
  publisher={Wiley Online Library}
}

@inproceedings{sonke2019adaptive,
  title={Adaptive radiotherapy for anatomical changes},
  author={Sonke, Jan-Jakob and Aznar, Marianne and Rasch, Coen},
  booktitle={Seminars in radiation oncology},
  volume={29},
  number={3},
  pages={245--257},
  year={2019},
  organization={Elsevier}
}

@article{sct1,
  title={Deep learning based synthetic-CT generation in radiotherapy and PET: a review},
  author={Spadea, Maria Francesca and Maspero, Matteo and Zaffino, Paolo and Seco, Joao},
  journal={Medical physics},
  volume={48},
  number={11},
  pages={6537--6566},
  year={2021},
  publisher={Wiley Online Library}
}

@article{gregoire2011state,
  title={State of the art on dose prescription, reporting and recording in Intensity-Modulated Radiation Therapy (ICRU report No. 83)},
  author={Gr{\'e}goire, Vincent and Mackie, Thomas R},
  journal={Cancer/radioth{\'e}rapie},
  volume={15},
  number={6-7},
  pages={555--559},
  year={2011},
  publisher={Elsevier}
}

@article{guo2025deep,
  title={Deep Learning for CT Synthesis in Radiotherapy},
  author={Guo, Yike and Luo, Yi and Hooshangnejad, Hamed and Zhang, Rui and Feng, Xue and Chen, Quan and Ngwa, Wilfred and Ding, Kai},
  journal={Bioengineering},
  volume={12},
  number={12},
  pages={1297},
  year={2025},
  publisher={MDPI}
}

@article{isensee2021nnu,
  title={nnU-Net: a self-configuring method for deep learning-based biomedical image segmentation},
  author={Isensee, Fabian and Jaeger, Paul F and Kohl, Simon AA and Petersen, Jens and Maier-Hein, Klaus H},
  journal={Nature methods},
  volume={18},
  number={2},
  pages={203--211},
  year={2021},
  publisher={Nature Publishing Group}
}

@article{siebert2024convexadam,
  title={Convexadam: Self-configuring dual-optimisation-based 3d multitask medical image registration},
  author={Siebert, Hanna and Gro{\ss}br{\"o}hmer, Christoph and Hansen, Lasse and Heinrich, Mattias P},
  journal={IEEE Transactions on Medical Imaging},
  year={2024},
  publisher={IEEE}
}

@article{fuxin,
  title={Enhancing Cross-Modality Synthesis: Subvolume Merging for MRI-to-CT Conversion},
  author={Fan, Fuxin and Qiu, Jingna and Huang, Yixing and Maier, Andreas},
  journal={arXiv preprint arXiv:2409.05982},
  year={2024}
}

@inproceedings{roy2023mednext,
  title={Mednext: transformer-driven scaling of convnets for medical image segmentation},
  author={Roy, Saikat and Koehler, Gregor and Ulrich, Constantin and Baumgartner, Michael and Petersen, Jens and Isensee, Fabian and Jaeger, Paul F and Maier-Hein, Klaus H},
  booktitle={International Conference on Medical Image Computing and Computer-Assisted Intervention},
  pages={405--415},
  year={2023},
  organization={Springer}
}

@inproceedings{hatamizadeh2021swin,
  title={Swin unetr: Swin transformers for semantic segmentation of brain tumors in mri images},
  author={Hatamizadeh, Ali and Nath, Vishwesh and Tang, Yucheng and Yang, Dong and Roth, Holger R and Xu, Daguang},
  booktitle={International MICCAI brainlesion workshop},
  pages={272--284},
  year={2021},
  organization={Springer}
}

@inproceedings{pix2pix,
  title={Image-to-image translation with conditional adversarial networks},
  author={Isola, Phillip and Zhu, Jun-Yan and Zhou, Tinghui and Efros, Alexei A},
  booktitle={Proceedings of the IEEE conference on computer vision and pattern recognition},
  pages={1125--1134},
  year={2017}
}

@article{paszke2019pytorch,
  title={Pytorch: An imperative style, high-performance deep learning library},
  author={Paszke, Adam and Gross, Sam and Massa, Francisco and Lerer, Adam and Bradbury, James and Chanan, Gregory and Killeen, Trevor and Lin, Zeming and Gimelshein, Natalia and Antiga, Luca and others},
  journal={Advances in neural information processing systems},
  volume={32},
  year={2019}
}

@article{yu2019eagan,
  title={Ea-GANs: edge-aware generative adversarial networks for cross-modality MR image synthesis},
  author={Yu, Biting and Zhou, Luping and Wang, Lei and Shi, Yinghuan and Fripp, Jurgen and Bourgeat, Pierrick},
  journal={IEEE transactions on medical imaging},
  volume={38},
  number={7},
  pages={1750--1762},
  year={2019},
  publisher={IEEE}
}

@inproceedings{ronneberger2015u,
  title={U-net: Convolutional networks for biomedical image segmentation},
  author={Ronneberger, Olaf and Fischer, Philipp and Brox, Thomas},
  booktitle={International Conference on Medical image computing and computer-assisted intervention},
  pages={234--241},
  year={2015},
  organization={Springer}
}

@article{thummerer2021clinical,
  title={Clinical suitability of deep learning based synthetic CTs for adaptive proton therapy of lung cancer},
  author={Thummerer, Adrian and Seller Oria, Carmen and Zaffino, Paolo and Meijers, Arturs and Guterres Marmitt, Gabriel and Wijsman, Robin and Seco, Joao and Langendijk, Johannes Albertus and Knopf, Antje-Christin and Spadea, Maria Francesca and others},
  journal={Medical physics},
  volume={48},
  number={12},
  pages={7673--7684},
  year={2021},
  publisher={Wiley Online Library}
}

@article{yoganathan2023generating,
  title={Generating synthetic images from cone beam computed tomography using self-attention residual UNet for head and neck radiotherapy},
  author={Yoganathan, SA and Aouadi, Souha and Ahmed, Sharib and Paloor, Satheesh and Torfeh, Tarraf and Al-Hammadi, Noora and Hammoud, Rabih},
  journal={Physics and Imaging in Radiation Oncology},
  volume={28},
  pages={100512},
  year={2023},
  publisher={Elsevier}
}

@inproceedings{zhu2017unpaired,
  title={Unpaired image-to-image translation using cycle-consistent adversarial networks},
  author={Zhu, Jun-Yan and Park, Taesung and Isola, Phillip and Efros, Alexei A},
  booktitle={Proceedings of the IEEE international conference on computer vision},
  pages={2223--2232},
  year={2017}
}

@article{peng2024cbct,
  title={CBCT-Based synthetic CT image generation using conditional denoising diffusion probabilistic model},
  author={Peng, Junbo and Qiu, Richard LJ and Wynne, Jacob F and Chang, Chih-Wei and Pan, Shaoyan and Wang, Tonghe and Roper, Justin and Liu, Tian and Patel, Pretesh R and Yu, David S and others},
  journal={Medical physics},
  volume={51},
  number={3},
  pages={1847--1859},
  year={2024},
  publisher={Wiley Online Library}
}

@article{gong2025boundary,
  title={Boundary information-guided adversarial diffusion model for efficient unsupervised synthetic CT generation},
  author={Gong, Changfei and Jian, Junming and Huang, Yuling and Luo, Mingming and Ding, Shenggou and Yuan, Xingxing and Wang, Xiaoping and Zhang, Yun},
  journal={Medical Physics},
  year={2025},
  publisher={Wiley Online Library}
}

@article{poch2025segmentation,
  title={Segmentation-Guided CT Synthesis with Pixel-Wise Conformal Uncertainty Bounds},
  author={Poch, David Vallmanya and Estievenart, Yorick and Zhalieva, Elnura and Patra, Sukanya and Yaqub, Mohammad and Taieb, Souhaib Ben},
  journal={arXiv preprint arXiv:2503.08515},
  year={2025}
}

@online{murphy2017windowingct,
  author  = {Murphy, Andrew and Feger, J. and Ismail, M. and others},
  title   = {Windowing (CT)},
  year    = {2017},
  date    = {2017-03-22},
  doi     = {10.53347/rID-52108},
  url     = {https://radiopaedia.org/articles/52108},
  urldate = {2026-02-09},
  note    = {Last revised: 2025-01-07}
}

@inproceedings{phan2024structural,
  title={Structural attention: Rethinking transformer for unpaired medical image synthesis},
  author={Phan, Vu Minh Hieu and Xie, Yutong and Zhang, Bowen and Qi, Yuankai and Liao, Zhibin and Perperidis, Antonios and Phung, Son Lam and Verjans, Johan W and To, Minh-Son},
  booktitle={International Conference on Medical Image Computing and Computer-Assisted Intervention},
  pages={690--700},
  year={2024},
  organization={Springer}
}

@article{ma2025vision,
  title={A vision--language pretrained transformer for versatile clinical respiratory disease applications},
  author={Ma, Liangdi and Liang, Hengrui and He, Yuwei and Wang, Wei and Yan, Zeping and Li, Wuchao and Wang, Rongpin and Li, Yongyi and Lizhu, Yuerong and Liu, Yaou and others},
  journal={Nature Biomedical Engineering},
  pages={1--19},
  year={2025},
  publisher={Nature Publishing Group UK London}
}

@article{howard2017mobilenets,
  title   = {MobileNets: Efficient Convolutional Neural Networks for Mobile Vision Applications},
  author  = {Howard, Andrew G. and Zhu, Menglong and Chen, Bo and Kalenichenko, Dmitry and Wang, Weijun and Weyand, Tobias and Andreetto, Marco and Adam, Hartwig},
  journal = {arXiv preprint arXiv:1704.04861},
  year    = {2017}
}

@article{mei2025ganeXt,
  title        = {GANeXt: A Fully ConvNeXt-Enhanced Generative Adversarial Network for MRI- and CBCT-to-CT Synthesis},
  author       = {Mei, Siyuan and Xia, Yan and Fan, Fuxin},
  journal      = {arXiv preprint arXiv:2512.19336},
  year         = {2025},
  archivePrefix= {arXiv},
  eprint       = {2512.19336},
  primaryClass = {cs.CV}
}

@inproceedings{yu2024inceptionnext,
  title={Inceptionnext: When inception meets convnext},
  author={Yu, Weihao and Zhou, Pan and Yan, Shuicheng and Wang, Xinchao},
  booktitle={Proceedings of the IEEE/cvf conference on computer vision and pattern recognition},
  pages={5672--5683},
  year={2024}
}

@article{vaswani2017attention,
  title={Attention is all you need},
  author={Vaswani, Ashish and Shazeer, Noam and Parmar, Niki and Uszkoreit, Jakob and Jones, Llion and Gomez, Aidan N and Kaiser, {\L}ukasz and Polosukhin, Illia},
  journal={Advances in neural information processing systems},
  volume={30},
  year={2017}
}

@inproceedings{lpips,
  title={The unreasonable effectiveness of deep features as a perceptual metric},
  author={Zhang, Richard and Isola, Phillip and Efros, Alexei A and Shechtman, Eli and Wang, Oliver},
  booktitle={Proceedings of the IEEE conference on computer vision and pattern recognition},
  pages={586--595},
  year={2018}
}





% LNCS Homepage, \url{http://www.springer.com/lncs}, last accessed 2023/10/25
% \end{thebibliography}
\end{document}